\documentclass[11pt,a4paper]{article}
\usepackage[hyperref]{emnlp2020}
\usepackage{times}
\usepackage{latexsym}

\usepackage{multirow}
\usepackage{natbib}
\usepackage{graphicx}
\usepackage{booktabs}
\usepackage{microtype}
\usepackage{caption}
\usepackage{cite}
\aclfinalcopy 

\title{Fancy Man Lauches Zippo at WNUT 2020 Shared Task-1: A Bert Case Model for Wet Lab Entity Extraction}

\author{Haoding Meng \\
  Xian Jiaotong University \\
  \small \texttt{menghd@stu.xjtu.edu.cn}
  \And
  Qingcheng Zeng \\
   The University of Manchester\\
  \small \texttt{qingcheng.zeng@student.manchester.ac.uk}
  \AND
  Xiaoyang Fang \\
 East China University of Science and Technology\\
 \small  \texttt{10182412@mail.ecust.edu.cn}
  \And
  Zhexin Liang \\
  Zhejiang University\\
  \small \texttt{3170103561@zju.edu.cn}}

\date{}
\begin{document}
\maketitle
\begin{abstract}
Automatic or semi-automatic conversion of protocols specifying steps in performing a lab procedure into machine-readable format benefits biological research a lot. These noisy, dense, and domain-specific lab protocols processing draws more and more interests with the development of deep learning. This paper presents our teamwork on WNUT 2020 shared task-1: wet lab entity extract, that we conducted studies in several models, including a BiLSTM CRF model and a Bert case model which can be used to complete wet lab entity extraction. And we mainly discussed the performance differences of \textbf{Bert case} under different situations such as \emph{transformers} versions, case sensitivity that may don't get enough attention before.
\end{abstract}

\section{Introduction}
The task of named entity recognition (NER) was first put forward in \citeyear{rau1991extracting}, after which it gradually became an essential part of natural language processing (NLP). The methods for NER is generally classified into four kinds: rule-based approaches, unsupervised learning approaches, feature-based supervised learning approaches and deep-learning based approaches. The previous methods are mainly rule-based, performing well on small dataset, like the LaSIE-II system provided by \citeauthor{Humphreys1995University}. Under the rapid development of deep learning since 2013, methods like \textbf{BiLSTM CRF} has been a hot spot in recent years. Even now, most of the deep learning methods for NER are based on this framework. But lately, some new research based on the concept of "pre-training" has attracted more and more attention, for example, Bert, which stands for bidirectional encoder representations from transformers \citep{2018BERT}. It both pushed the GLUE score to 80.5 \% , which got 7.7 \% point absolute improvement and created a new paradigm of natural language processing task method: using the model pre-trained on a large corpus to complete downstream tasks through fine-tuning.

\begin{figure}
\fbox{%
\small
\parbox{0.45\textwidth}{
\textbf{Isolation of temperate phages by plaque agar overlay}\\
1. Melt soft agar overlay tubes in boiling water and place in the 47C water bath.\\
2. Remove one tube of soft agar from the water bath.\\
3. Add 1.0 mL host culture and either 1.0 or 0.1 mL viral concentrate.\\
4. Mix the contents of the tube well by rolling back and forth between two hands, and immediately empty the tube contents onto an agar plate.\\
5. Sit RT for 5 min.\\
6. Gently spread the top agar over the agar surface by sliding the plate on the bench surface using a circular motion.\\
7. Harden the top agar by not disturbing the plates for 30 min.\\
8. Incubate the plates (top agar side down) overnight to 48h.\\
9. Temperate phage plaques will appear as turbid or cloudy plaques, whereas purely lytic phage will appear as shaply defined, clear plaques.}
}
\caption{An example wet lab protocol \citep{kulkarni2018an}}
\label{fig:examp}
\end{figure}

\section{Task}
\begin{figure}
    \centering
    \includegraphics[width=3.0in]{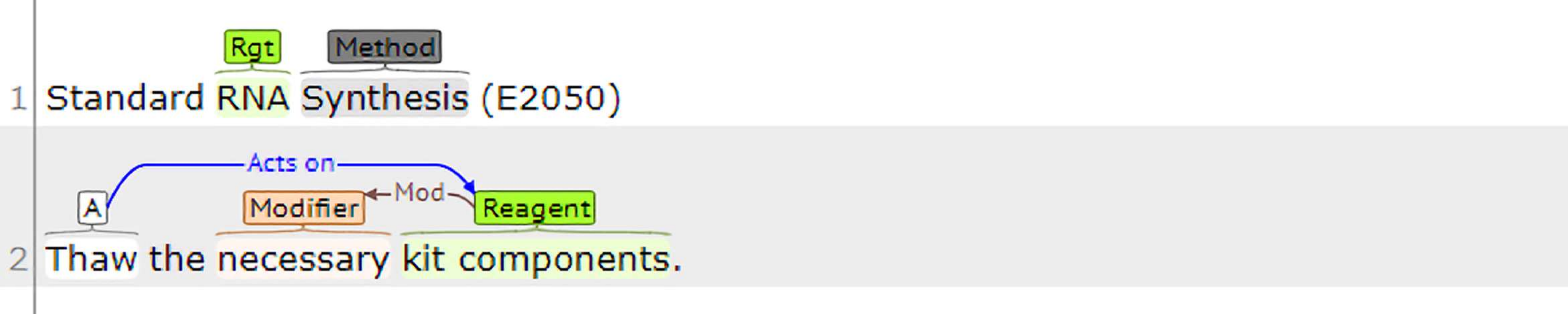}
    \caption{A visualization of the BRAT style annotation}
    \label{fig:2}
\end{figure}

For WNUT 2020 shared task-1 \citep{tabassum2020wlp}, participants were asked to develop a system that automatically identify entities from the provided lab instructions dataset. The dataset is from wet lab protocols, which usually refer to the experiment instructions in biology or chemistry experiments, involving substances like chemicals, proteins, drugs and other materials. Figure \ref{fig:examp} shows one representative example of the wet lab protocols.
In this shared task, the data was divided into three parts, training data with 370 protocols, development data with 122 protocols and test data with 123 protocols. The data was given in CoNLL format. To sum up, this was a small dataset in size, and specialized in laboratory settings.
The dataset is annotated by the researchers in Ohio State University \citep{kulkarni2018an} with BRAT \citep{stenetorp2012brat:}. And it could be visualized via \url{http://bit.ly/WNUT2020 platform}. Figure \ref{fig:2} shows a visualization result of the protocol 3 in our training dataset.

In this dataset, there are 18 kinds of entities, as shown in table \ref{tab:type}.

\begin{table}[]
\scalebox{0.98}{%
		\begin{tabular}{cll}
			\toprule
			& \textbf{Type}            & \textbf{Example}              \\ \midrule
			1  & Method          & Extraction           \\
			2  & Modifier        & High Quality Genomic \\
			3  & Reagent         & DNA                  \\
			4  & Action          & dissect              \\
			5  & Amount          & 1-10mg               \\
			6  & Device          & Flow Cytometer       \\
			7  & Time            & 5 minutes            \\
			8  & Speed           &350xg                \\
			9  & Mention         & ethanol wash         \\
			10 & Location        & tube                 \\
			11 & Numerical       & 10 times            \\
			12 & Temperature     & 60C                  \\
			13 & Size            & 0.45m               \\
			14 & Concentration   & 4\%                 \\
			15 & Measure-Type    & volume               \\
			16 & Generic-Measure & TFSC=40             \\
			17 & Seal            & bottle cap           \\
			18 & pH              & pH 8.0               \\ \bottomrule
		\end{tabular}%
	}
	\caption{Named entity types in WNUT 2020 shared task-1}
	\label{tab:type}
\end{table}

\section{Model}

\subsection{Baseline}
\label{sec:baseline}
The provided baseline model is a linear conditional random field (CRF) tagger, which is one of the traditional machine learning ways to complete the named entity recognition task \citep{10.3115/1219840.1219885}. This tagger does the NER task with feature engineering, taking word features, context features and gazatteer features into consideration. 

\subsection{BiLSTM CRF}
\label{sec:bilstm crf}
\textbf{BiLSTM CRF} is a deep learning oriented tagger to complete the NER task \citep{huang2015bidirectional}. The long-short term memory (LSTM) unit \citep{1997Long} is a kind of specifically designed recurrent neural network (RNN) to process the timing sequential information. Here, LSTM units are adopted to collect the information in the context. Additionally, they are bidirectional so that they can take information from both sides into consideration.

One more CRF layer is added into this model because it will help the model to standardizing  output results. For example, a sequence like “B-Action I-Mention” will never be possible in real world. However, a pure BiLSTM model is possibly giving this kind of errors. 

\begin{figure}[htp]
    \centering
    \includegraphics[width=3.0in]{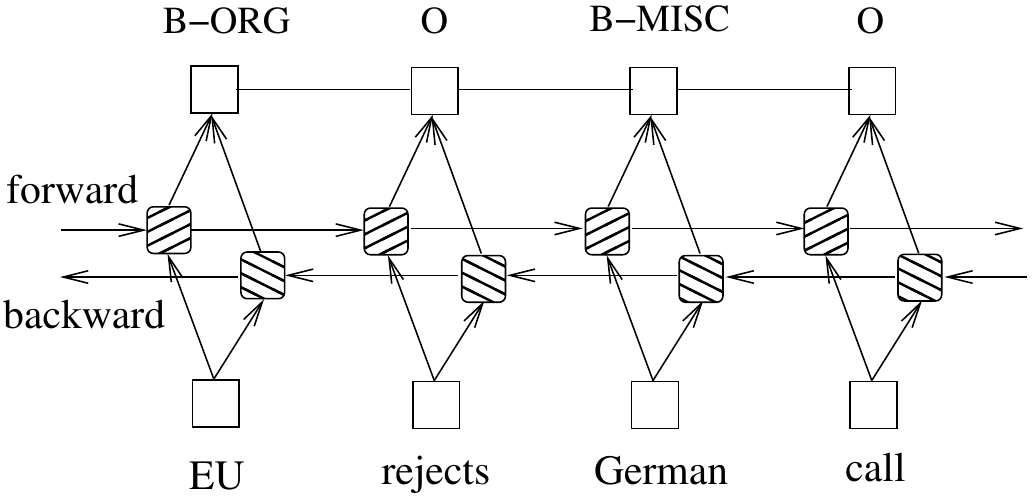}
    \caption{The architecture of a BiLSTM CRF model \citep{huang2015bidirectional}}
    \label{fig:4}
\end{figure}

The basic architecture of \textbf{BiLSTM CRF} is shown in figure \ref{fig:4}.

\subsection{Bert}
\label{sect:bert}
Bidirectional encoder representations from transformers (Bert) was first proposed by Google AI researchers in 2018 \citep{2018BERT}. It achieved quite a few new records in NLP field and the concept of “pre-training” has been popular since then.

In this shared task, we also adopted Bert pre-training model to do the NER task and to compare the results with \textbf{BiLSTM CRF} to explore the performance of different techniques.

Our \textbf{Bert case} adopted \emph{BertforTokenClassification} class in \emph{transformers} \citep{2019HuggingFace} and added one more fine-tuning layer to complete the task. The architecture is shown in the following figure \ref{fig:5}.

\begin{figure}[htp]
    \centering
    \includegraphics[width=3.0in]{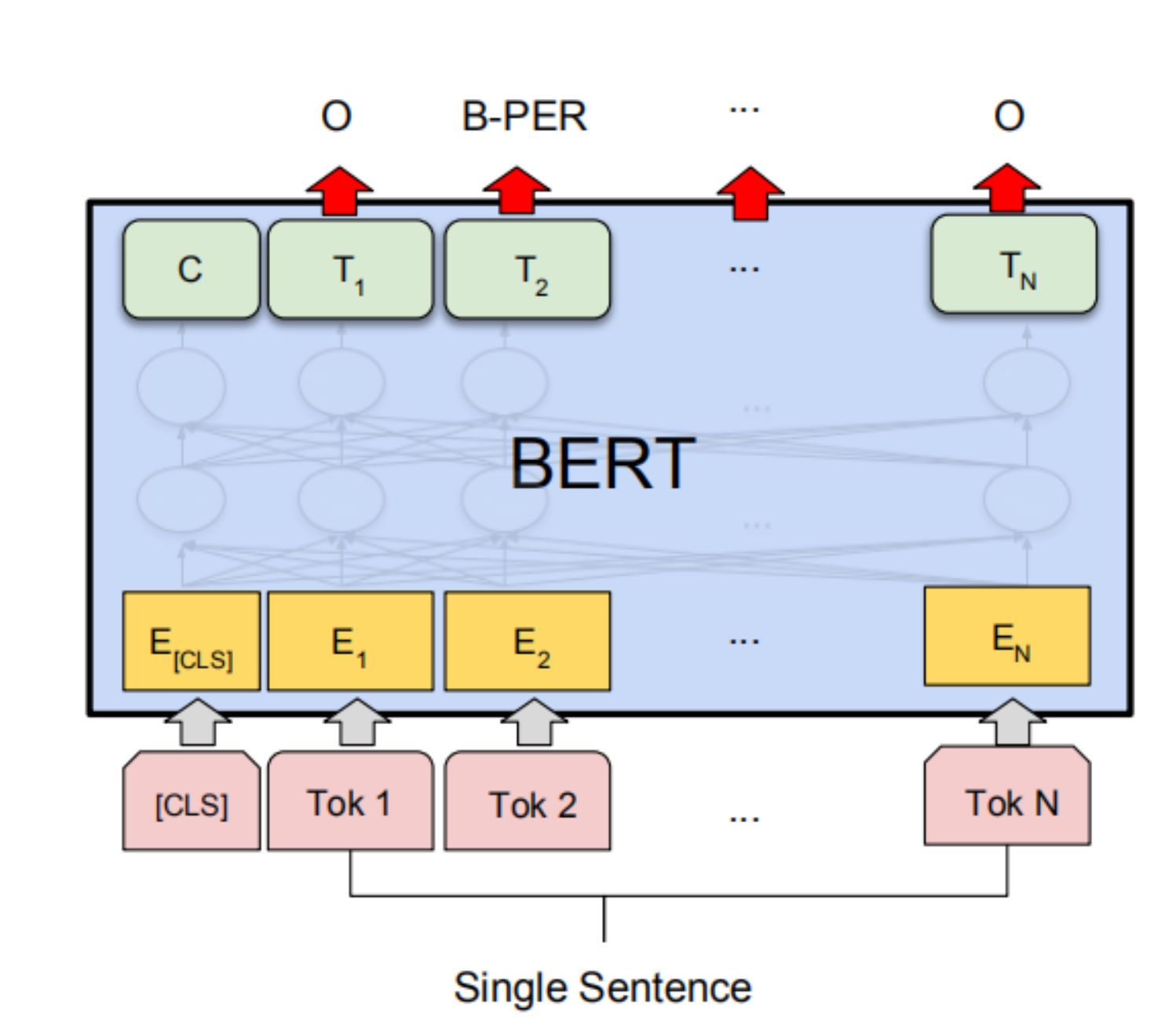}
    \caption{Bert fine-tuning for token classification \citep{2018BERT}}
    \label{fig:5}
\end{figure}

\section{Experiment}
In this shared task, we tried to compare the relatively traditional deep learning method, \textbf{BiLSTM CRF}, and pre-training Bert based, including Bert base cased model and Bert base uncased model. The former needs to train static word vector from dataset, while the latter is equivalent to using dynamic word vector and we can directly conduct fine-tuning experiments on NER by connecting \emph{BertforTokenClassification} after pre-training model. It can be seen that from table \ref{tab:bilstm} that the performance of \textbf{BiLSTM CRF} without careful training of word vector is not as good as the baseline model provided, \textbf{Linear CRF}. Therefore, our follow-up experiments will focus on training Bert in different situations and carry out exploration and discussion.

\begin{table}[htbp]
\scalebox{1.0}{%
\begin{tabular}{lccc}
\toprule
           & precision & recall & F1     \\ \midrule
\textbf{Linear CRF} & 0.7549    & 0.7332 & 0.7439 \\
\textbf{BiLSTM CRF} & 0.7208    & 0.6605 & 0.7101 \\ \bottomrule
\end{tabular}%
}
\caption{Results of \textbf{BiLSTM CRF} and baseline}
\label{tab:bilstm}
\end{table}

\subsection{Stipulate}
For the sake of simplicity, we will use $wd$ and $\eta$ represent weight decay and learning rate respectively and \textbf{cased/uncased with/without} to abbreviate the corresponding model with lowercase processing or not in subsequent trials. In addition, \emph{v} means importing the required classes like \emph{BertModel} from \emph{pytorch-transformers}, and \emph{V}  means importing from \emph{transformers}, and classes are imported from the latter by default. In general, the precision and recall of the model are considered comprehensively in f1-score, so the performance of the model is often evaluated using f1-score. And the numbers \underline{underlined} in the chart represent possible anomalies, while the numbers highlighted in \textbf{bold} represent the best results.

The model doesn't converge when it's trained only once, but it may face problems of over fitting and CUDA out of memory if it is more than 4, for only 8g memory in RTX 2060 Super and RTX 2080. After several trials, 3 is selected as the optimal default epoch number. For alleviating over fitting, weight decay technique (or L2 regularization) is usually adopted and $wd$ is empirically set between 0.001 and 0.01. If there is no special explanation, the default value of weight decay value in this paper is 0.005.
\subsection{Train}

\subsubsection{Learning Rate}
Named entity recognition is one of downstream tasks of Bert. Since it has been trained on a large scale corpus, the recommended learning rate is generally small, such as $2e-5$, $3e-5$, $5e-5$ \citep{2018BERT}. But this needs to be considering with the specific application scenarios. By using the default Bert model: \textbf{cased without} and \textbf{uncased with}, we trained on both 2060s and 2080 with different learning rates. 
\begin{table*}[htbp]
\resizebox{\textwidth}{!}{%
\begin{tabular}{llcccccccccc}
\toprule
\multicolumn{2}{c}{learning rate} & 2e-5   & 3e-5   & 5e-5   & 8e-5   & 9e-5   & 1e-4   & 2e-4                        & 3e-4   & 4e-4   & 5e-4   \\ \midrule
          & \textbf{cased without}         & 0.7776 & 0.7848 & 0.7909 & 0.7949 & 0.7961 & 0.7965 & \textbf{0.7980} & 0.7963 & 0.7956 & 0.7911 \\
\multirow{-2}{*}{2060s} &
  \textbf{uncased with} &
  0.7797 &
  0.7900 &
  0.7974 &
  0.7987 &
  \textbf{0.7993} &
  \textbf{0.7993} &
  0.7972 &
  0.7968 &
  0.7970 &
  0.7928 \\
          & \textbf{cased without}         & 0.7775 & 0.7844 & 0.7907 & 0.7951 & 0.7973 & 0.7994 & \textbf{0.8008} & 0.7946 & 0.7917 & 0.7876 \\
\multirow{-2}{*}{2080} &
  \textbf{uncased with} &
  0.7775 &
  0.7881 &
  0.7948 &
  0.7974 &
  0.7956 &
  0.7962 &
  \textbf{0.7988} &
  0.7948 &
  0.7974 &
  0.7903 \\ \bottomrule
\end{tabular}%
}
\caption{Bert case default model performances at differing learning rates}
\label{tab:lr}
\end{table*}

As shown in table \ref{tab:lr}, all recommended learning rates did not perform well in this task, and we thought that the dataset provided this time are not common in daily life, so we were ought to increase the learning rate appropriately. 
At the same time, we noticed that the uncased model is better than the cased model on 2060s, but it is opposite in 2080. Although the best model is obtained by training on 2080, the model training on 2060s is more stable.

\subsubsection{Case Sensitivity and Version}
Generally speaking, the uncased model is better than the cased model, however, the cased model performs better when there are obvious case differences in specific aspects such as named entity recognition. But we also noticed that we could train an uncased model after processing the text in lowercase.

During testing, we also found that different versions of classes imported will lead to differing results. To better explore the influence of case sensitivity and version, we further trained three possible casing methods, including: \textbf{cased without}, \textbf{cased with} and \textbf{uncased with} (by the way, \textbf{uncased without} should perform the worst, because it can't actually distinguish case information, and the experimental results are exactly the same, so we omit this possible combination) and two different versions of the combination model. Record the performance of each model and take the top two to get table \ref{tab:library}.

Except for some models with slight performance decrease, in most cases, the model can be further improved by using \emph{pytorch-transformers} to import the required classes. Moreover, we noticed that \textbf{uncased with} is the best model on both 2060s and 2080 when considering the use of previous versions of classes (\textbf{uncased with} \emph{V}), however, when only the latest version of \emph{transformers} is used, the cased model works best. But the relationship between whether to use lowercase processing and the final performance is not obvious from our experiments.

 Theoretically, words with different case could represent the same named entity, while using lowercase processing can increase the number of training samples but reduce the number of types. So we suggest that when using the updated \emph{transformers} training, please use \textbf{cased without}, and when using the previous version training, consider using \textbf{uncased with}. What's more, \textbf{cased with} is also an option worth considering.

\subsubsection{Weight Decay}
\begin{figure}[htp]
    \centering
    \includegraphics[width=3.0in]{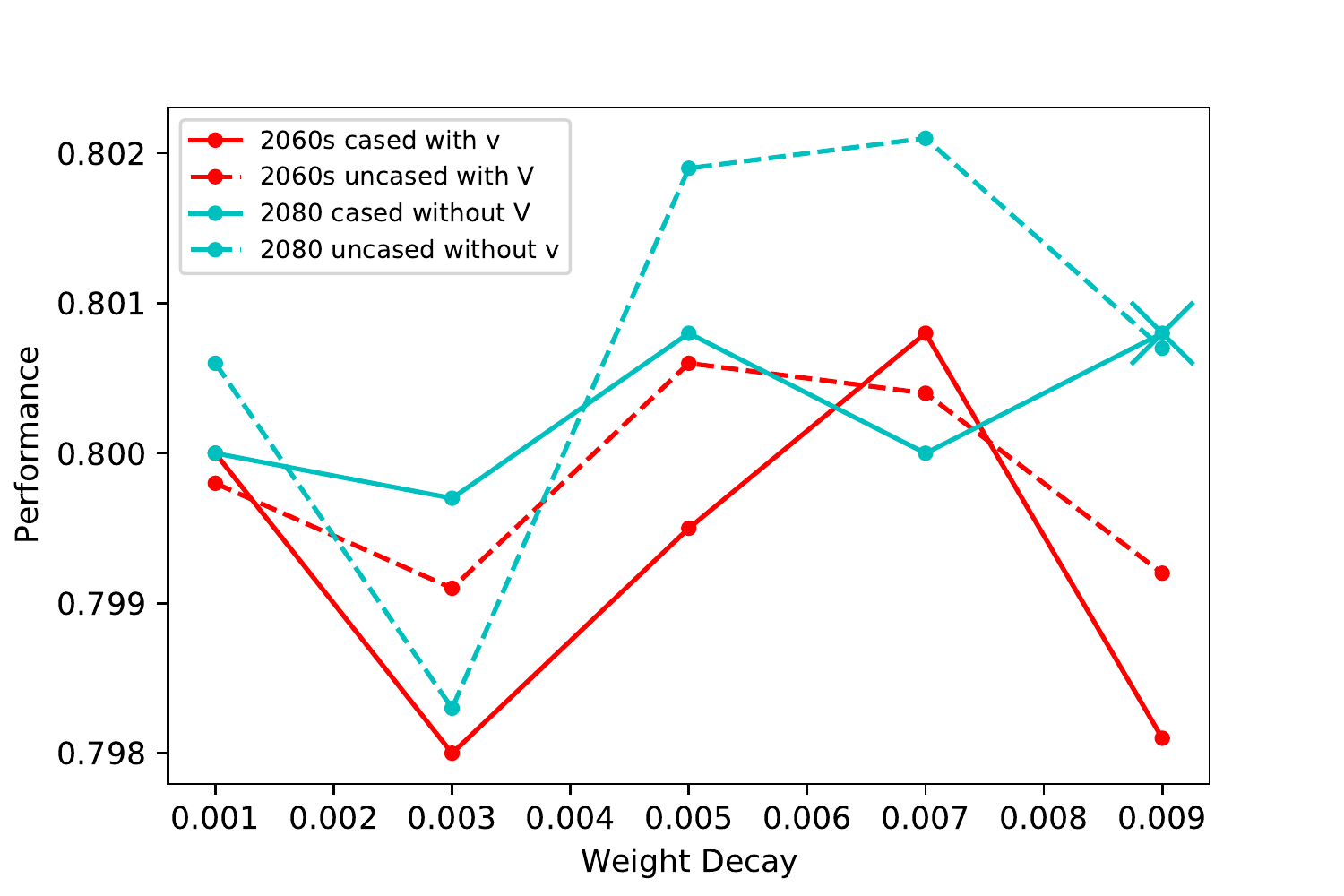}
    \caption{The tendency comparisons under different weight decay and experimental conditions}
    \label{fig:6}
\end{figure}

We used 0.005 as our default weight decay value before, which based on several simple attempts. Here, we selected the best four models on 2060s and 2080 respectively to adjust the weight decay under the condition of using previous version classes or not, and sorted them out as figure \ref{fig:6}.

In theory, the increase of weight decay will make the performance of the model increase to the maximum firstly and then decrease. This is because at the beginning, the model performs poorly in the test set due to over fitting. With the increase of penalty term, the performance of the model is improved and the best value is generated. If the penalty term continues to increase, the model will tend to be more simple model, so the performance of the model drops. However, the actual situation in the evaluation is that the performance of the model firstly increases and then decreases with the increase of weight decay, and then increases to the maximum value and then decreases according to the theory. It is worth noting that the f1-score of \textbf{case without} when weight decay value is 0.009 is the same as that is 0.005, which is difficult to explain according to the classical theory (more specifically, the recall of the former is higher, but the precision of the latter is higher). Though it's true that the best performance of most models is achieved at 0.005, the final selected model in this experiment is \textbf{uncased with} \emph{V} on 2080 when the weight decay value is 0.007. The specific index comparison table between this Base case model and baseline is shown in table \ref{tab:vs}.

\textbf{Bert case} based on Transformer has achieved better results in the recognition of most entity categories than \textbf{Linear CRF} combined with traditional feature engineering. It can be seen that using a large corpus for pre-training combined with downstream task fine-tuning strategies is a very effective and operational relatively simple model paradigm. However, for the output results of \textbf{Bert case} without CRF standardizing, we can find that many unreasonable annotation results are generated in the test set, and on some indicators, the model does not perform as well as the baseline, especially in the prediction results on Temperature and Measure. All in all, Bert can be used as a relatively good benchmark, but there is still much room for improvement. We have planned to replace it with other pre-trained models such as Roberta, Albert, XLNet, and connected it with CRF to observe the effect of the experiment, but it finally failed to achieve due to limited time and capacity.

\begin{table}[htbp]
\scalebox{0.53}{%
\begin{tabular}{clllllll}
\toprule
                                                  &                           & \multicolumn{2}{c}{\textbf{cased without}} & \multicolumn{2}{c}{\textbf{uncased with}} & \multicolumn{2}{c}{\textbf{cased with}} \\ \midrule
\multicolumn{1}{c}{\multirow{2}{*}{\emph{transformers}}} & 2080                      & 0.8008          & 0.7994          & 0.7988         & 0.7974          & 0.7979         & 0.7956        \\
\multicolumn{1}{c}{}                              & \multicolumn{1}{r}{2060s} & 0.7980           & 0.7965          & 0.7993         & 0.7993          & 0.7995         & 0.7960         \\
\multirow{2}{*}{\emph{pytorch-transformers}}             & 2080                      & 0.8004          & 0.7997          & 0.8019         & 0.7992          & 0.7975         & 0.7975        \\
                                                  & \multicolumn{1}{r}{2060s} & 0.7990           & 0.7982          & 0.8006         & 0.7994          & 0.7993         & 0.7997        \\ \midrule
\textbf{avg}                        &                           & +0.03\%          & +0.12\%           & +0.28\%   & +0.11\%         & -0.03\%        & +0.35\%        \\ \bottomrule
\end{tabular}%
}
\caption{The performance change under different versions and case sensitivity}
\label{tab:library}
\end{table}

\subsection{Test}
The classification results of each group participating in this task can be accessed from \href{https://docs.google.com/spreadsheets/d/1Iu0FfkradDgrMc9KTvByNrUyMSaVT1JFkCYXpt4oZRA/edit?usp=sharing}{results}. As the final test model was submitted before, it could not be trained from aspects proposed in this paper, and the model could only be trained on RTX 2060 Super, so the final model is \textbf{uncased with} with lower f1-score. In addition to these influencing factors shown above, we also noted that on different operating systems (Ubuntu and Windows), whether or not to enable X service and perform other tasks during training may also change the performance of the model with the same other conditions. However, due to its complexity, we have not got the results temporarily and we hope further research could carry on in the future.

\begin{table}[htbp]
\scalebox{0.65}{%
\begin{tabular}{lcccccc}
\toprule
              & \multicolumn{3}{c}{\textbf{Linear CRF}} & \multicolumn{3}{c}{\textbf{Bert case}}              \\
              & precision  & recall  & F1      & precision    & recall       & F1           \\ \midrule
Action        & 0.8456     & 0.8423  & 0.8440  & 0.8889       & 0.9179       & 0.9031       \\
Amount        & 0.8521     & 0.8319  & 0.8419  & 0.8798       & 0.9078       & 0.8936       \\
Concentration & 0.7770     & 0.7929  & 0.7849  & 0.8230       & 0.8659       & 0.8439       \\
Device        & 0.6125     & 0.5629  & 0.5867  & 0.6455       & 0.6966       & 0.6701       \\
Measure       & 0.3600     & 0.2553  & 0.2988  & \underline{0.3119} & \underline{0.2378} & \underline{0.2698} \\
Location      & 0.6883     & 0.6951  & 0.6917  & 0.7525       & 0.7882       & 0.7700       \\
Type          & 0.5164     & 0.4649  & 0.4893  & \underline{0.4713} & 0.5735       & 0.5174       \\
Mention       & 0.5965     & 0.6071  & 0.6018  & 0.6567       & 0.7857       & 0.7154       \\
Method        & 0.5069     & 0.3830  & 0.4363  & 0.5080       & 0.4914       & 0.4996       \\
Modifier      & 0.5682     & 0.5134  & 0.5394  & 0.5968       & 0.6140       & 0.6053       \\
Numerical     & 0.5636     & 0.5758  & 0.5696  & 0.6000       & 0.6623       & 0.6296       \\
Reagent       & 0.7475     & 0.7522  & 0.7498  & 0.8161       & 0.8412       & 0.8284       \\
Seal          & 0.6825     & 0.6719  & 0.6772  & \underline{0.6712} & 0.7656       & 0.7153       \\
Size          & 0.6667     & 0.5000  & 0.5714  & 0.7805       & 0.5614       & 0.6531       \\
Speed         & 0.8421     & 0.8675  & 0.8546  & 0.8908       & 0.9281       & 0.9091       \\
Temperature   & 0.9385     & 0.8975  & 0.9176  & \underline{0.9130} & 0.9079 & \underline{0.9105} \\
Time          & 0.8969     & 0.8762  & 0.8864  & \underline{0.8857}      & 0.9080       & 0.8967       \\
pH            & 0.7255     & 0.5968  & 0.6549  & 0.7273       & 0.7742       & 0.7500       \\ \midrule
\textbf{avg}           & 0.7549     & 0.7332  & 0.7439  & 0.7897       & 0.8148       & 0.8021       \\ \bottomrule
\end{tabular}%
}
\caption{Comparison of classification results between \textbf{Bert case} and \textbf{Linear CRF}}
\label{tab:vs}
\end{table}

\section{Conclusion}
This article introduces our relevant experimental research based on this WNUT 2020 shared task-1. By trying \textbf{BiLSTM CRF}, we learn about that the method based on static word vector needs to be trained on a specific dataset, so its transferability is relatively low. And we mainly focused on the fine-tuning experiments based on Bert under different conditions, including learning rate, GPU, \emph{transformers} version, case sensitivity and weight decay, and conducted discussions, so as to understand that the possible influencing factors in actual model training. It is quite necessary to unify and clarify experimental conditions while evaluating the performance of related models in the future, because even the class imported matters.

We noted that recently, several papers presented NER studies using convolutional neural network (CNN) \citep{li2018biomedical,zhai2018comparing}. This may imply that more work combining pre-trained model and CNN will be a new direction for NER studies. Additionally, although more and more attention has been focused on deep learning nowadays, the methods based on traditional machine learning still get attention and continue to develop with its interpretability and robustness in specific domain tasks.


\begin{thebibliography}{12}
\expandafter\ifx\csname natexlab\endcsname\relax\def\natexlab#1{#1}\fi

\bibitem[{Devlin et~al.(2018)Devlin, Chang, Lee, and Toutanova}]{2018BERT}
Jacob Devlin, Ming~Wei Chang, Kenton Lee, and Kristina Toutanova. 2018.
\newblock Bert: Pre-training of deep bidirectional transformers for language
  understanding.

\bibitem[{Finkel et~al.(2005)Finkel, Grenager, and
  Manning}]{10.3115/1219840.1219885}
Jenny~Rose Finkel, Trond Grenager, and Christopher Manning. 2005.
\newblock \href {https://doi.org/10.3115/1219840.1219885} {Incorporating
  non-local information into information extraction systems by gibbs sampling}.
\newblock In \emph{Proceedings of the 43rd Annual Meeting on Association for
  Computational Linguistics}, ACL '05, page 363–370, USA. Association for
  Computational Linguistics.

\bibitem[{Hochreiter and Schmidhuber(1997)}]{1997Long}
Sepp Hochreiter and JüRgen~A Schmidhuber. 1997.
\newblock Long short-term memory.
\newblock \emph{Neural Computation}.

\bibitem[{Huang et~al.(2015)Huang, Xu, and Yu}]{huang2015bidirectional}
Zhiheng Huang, Wei Xu, and Kai Yu. 2015.
\newblock Bidirectional lstm-crf models for sequence tagging.
\newblock \emph{arXiv: Computation and Language}.

\bibitem[{Humphreys et~al.(1995)Humphreys, Gaizauskas, Azzam, Huyck, and
  Wilks}]{Humphreys1995University}
K.~Humphreys, R.~Gaizauskas, S.~Azzam, C.~Huyck, and Y.~Wilks. 1995.
\newblock \emph{University of Sheffield: Description of the LaSIE-II system as
  used for MUC-7}.
\newblock Association for Computational Linguistics.

\bibitem[{Kulkarni et~al.(2018)Kulkarni, Xu, Ritter, and
  Machiraju}]{kulkarni2018an}
Chaitanya Kulkarni, Wei Xu, Alan Ritter, and Raghu Machiraju. 2018.
\newblock An annotated corpus for machine reading of instructions in wet lab
  protocols.
\newblock 2:97--106.

\bibitem[{Li and Guo(2018)}]{li2018biomedical}
SL~Li and YK~Guo. 2018.
\newblock Biomedical named entity recognition with cnn-blstm-crf [j].
\newblock \emph{Journal of chinese information processing}, 32(1):116--122.

\bibitem[{Rau(1991)}]{rau1991extracting}
Lisa~F Rau. 1991.
\newblock Extracting company names from text.
\newblock In \emph{Proceedings The Seventh IEEE Conference on Artificial
  Intelligence Application}, pages 29--30. IEEE Computer Society.

\bibitem[{Stenetorp et~al.(2012)Stenetorp, Pyysalo, Topic, Ohta, Ananiadou, and
  Tsujii}]{stenetorp2012brat:}
Pontus Stenetorp, Sampo Pyysalo, Goran Topic, Tomoko Ohta, Sophia Ananiadou,
  and Junichi Tsujii. 2012.
\newblock brat: a web-based tool for nlp-assisted text annotation.
\newblock pages 102--107.

\bibitem[{Tabassum et~al.(2020)Tabassum, Xu, and Ritter}]{tabassum2020wlp}
Jeniya Tabassum, Wei Xu, and Alan Ritter. 2020.
\newblock {WNUT-2020 Task 1: Extracting Entities and Relations from Wet Lab
  Protocols}.
\newblock In \emph{Proceedings of EMNLP 2020 Workshop on Noisy User-generated
  Text (WNUT)}.

\bibitem[{Wolf et~al.(2019)Wolf, Debut, Sanh, Chaumond, Delangue, Moi, Cistac,
  Rault, Louf, and Funtowicz}]{2019HuggingFace}
Thomas Wolf, Lysandre Debut, Victor Sanh, Julien Chaumond, Clement Delangue,
  Anthony Moi, Pierric Cistac, Tim Rault, Rémi Louf, and Morgan~and Funtowicz.
  2019.
\newblock Huggingface's transformers: State-of-the-art natural language
  processing.

\bibitem[{Zhai et~al.(2018)Zhai, Nguyen, and Verspoor}]{zhai2018comparing}
Zenan Zhai, Dat~Quoc Nguyen, and Karin Verspoor. 2018.
\newblock Comparing cnn and lstm character-level embeddings in bilstm-crf
  models for chemical and disease named entity recognition.
\newblock \emph{arXiv preprint arXiv:1808.08450}.

\end{thebibliography}
\end{document}